\documentclass[twocolumn]{article}
\usepackage[utf8]{inputenc} 
\usepackage{graphicx}       
\usepackage{hyperref}       
\usepackage{array}          
\usepackage{booktabs}       
\usepackage{fancyhdr}       
\usepackage{enumitem}       
\usepackage{authblk}        
\usepackage{footmisc}       
\usepackage[T1]{fontenc}
\usepackage{CJKutf8}             
\usepackage{amsmath, amssymb}
\usepackage{graphicx}
\usepackage{geometry}
\usepackage[utf8]{inputenc}      
\usepackage[T1]{fontenc}
\usepackage{CJKutf8}             
\usepackage{amsmath, amssymb}
\usepackage{graphicx}
\usepackage{hyperref}
\usepackage{geometry}
\DeclareUnicodeCharacter{3000}{\hspace{1em}}

\title{Integrated ensemble of BERT- and features-based models for \\
  authorship attribution in Japanese literary works}

\author[1]{Taisei Kanda}
\author[2,3]{Mingzhe Jin\thanks{Current address: 1-3 Tatara Miyakotani, Kyotanabe, Kyoto, Japan.}\thanks{Corresponding author: \texttt{singane@gmail.com}}}
\author[4]{Wataru Zaitsu}

\affil[1]{Graduate School of Culture and Information Science, Doshisha University}
\affil[2]{Research Center for Linguistic Ecology, Doshisha University, Kyoto, Japan}
\affil[3]{Institute of Interdisciplinary Research, Kyoto University of Advanced Science, Kyoto, Japan}
\affil[4]{Faculty of Psychology, Mejiro University, Tokyo, Japan}

\date{}

\begin{document}

\maketitle

\begin{abstract}
Traditionally, authorship attribution (AA) tasks relied on statistical data analysis and classification based on stylistic features extracted from texts. In recent years, pre-trained language models (PLMs) have attracted significant attention in text classification tasks. However, although they demonstrate excellent performance on large-scale short-text datasets, their effectiveness remains under-explored for small samples, particularly in AA tasks. Additionally, a key challenge is how to effectively leverage PLMs in conjunction with traditional feature-based methods to advance AA research. In this study, we aimed to significantly improve performance using an integrated integrative ensemble of traditional feature-based and modern PLM-based methods on an AA task in a small sample. For the experiment, we used two corpora of literary works to classify 10 authors each. The results indicate that BERT is effective, even for small-sample AA tasks. Both BERT-based and classifier ensembles outperformed their respective stand-alone models, and the integrated ensemble approach further improved the scores significantly. For the corpus that was not included in the pre-training data, the integrated ensemble improved the F1 score by approximately 14 points, compared to the best-performing single model. Our methodology provides a viable solution for the efficient use of the ever-expanding array of data processing tools in the foreseeable future.　
\end{abstract}

\hspace{0.5mm}

\hspace{-4mm}\textbf{Keywords:} Bidirectional encoder representations from transformers (BERT), Stylometric features, Authorship attribution (AA), Text classification, Integrated ensemble.

\section{Introduction}
Authorship attribution (AA) entails assigning authorship to texts with unknown authorship [1, 2, 3]. Various studies on AA have been conducted, tracing back more than 100 years. 

Mendenhall [4] counted the number of letters in each word used in a sentence, analyzed the relative frequency curves, and demonstrated that the curves varied among authors and could be a distinctive characteristic of each author. Mendenhall [5] also demonstrated that Shakespeare predominantly used four-letter words, whereas Bacon favored three-letter words. This finding challenged the theory that Bacon authored a series of satirical plays under the pseudonym "Shakespeare" to protest against the oppressive government. These features are referred to as stylistic features. They are distinctive techniques or devices that an author uses to create a particular effect in a text. These are woven throughout the work. 

Before the 1950s, word, sentence, and paragraph lengths, which were easy to quantify, were mainly used for statistical analysis. With the development of computational environments, many scholars have proposed extracting stylistic features from text based on aspects such as character, word, part-of-speech structure, grammar, and syntax [2, 6-9]. Stylistic features that are mechanically aggregated from texts contain a lot of noise. 

Stylistic features, subsequently referred to as features in this paper, are language-dependent. For example, Japanese or Chinese differ from Western languages in character forms, writing styles, lack of segmentation, and grammatical structures. One of the most evident ways in which Japanese or Chinese are unique is the character forms and the fact that they are not divided into words. Therefore, the elements that appear as features, such as characters, words, and phrases, also vary depending on the segmentation method. 

 AA was first performed in stylistic studies of literary works. Over time, it has also been applied to detect fake news, address authorship issues, identify plagiarism, and investigate matters in criminal and civil law [9, 10, 11]. With the rapid development of computer science, powerful machine learning classifiers and pre-trained language models (PLMs) are being consecutively developed, and the AA environment is changing rapidly. 

Several classifiers are now commonly used for text classification, including penalized logistic regression, support vector machine (SVM), random forest (RF), boosting methods (such as AdaBoost, XGBoost), neural network approaches, and PLMs such as bidirectional encoder representations from transformers (BERT) and its derivatives, RoBERTa and DeBERTa. However, it is difficult to determine the best model. Moreover, an important issue for end users is how to adapt such tools to their specific tasks. Against this backdrop, this study focused on significantly improving AA scores in small-scale samples and literary works by employing an integrated ensemble that combines traditional feature- and classifier-based methods with PLM-based methods.

The remainder of this paper is organized as follows. Section 2 presents a review of pertinent literature, identifies the limitations of previous studies, and outlines the research questions. Sections 3 and 4 present the experimental procedure and study results, respectively. Sections 5 and 6 are the discussion and conclusions, respectively. 

\section{Related research}
Multivariate data analysis methods that use features extracted manually or automatically from texts have been used in AA tasks since 1960. These methods include unsupervised techniques, such as principal component analysis, correspondence analysis, and clustering, as well as supervised techniques, such as linear and nonlinear discriminant analysis. Since the 1990s, neural networks [12, 13], SVM, RF, boosting classifiers have been employed [2, 3, 14, 15].

Jin and Murakami [14] demonstrated that RF is more effective than SVM for noisy data. They also analyzed the decrease in the RF and SVM scores with decreasing sample size. Liu and Jin [15] conducted a comparative analysis of the genre and author classifications of mixed-genre texts using 14 feature datasets (character $n$-gram, $n=1, 2, 3$; POS tag $n$-gram, $n=1, 2, 3$; token $n$-gram, $n=1, 2, 3$; token and POS tag n-gram, $n=1, 2, 3$; phrase pattern; comma position ) and seven classifiers (SVM, RF, AdaBoost, HDDA, LMT, XGBoost, and Lasso). The results indicate that features and classifier scores vary across different cases. In other words, even within the same corpus, variations arise where combinations such as character bigrams with the RF classifier demonstrate higher score, whereas token unigrams paired with the AdaBoost classifier may show the best performance. 

Regarding the use of structured data features and classifiers in AA, there have been studies on methods for feature extraction and classifier arrangements; however, there has been no significant progress in score improvement. Considering this, Jin [16] proposed an ensemble approach that employed multiple feature vectors (character bigram, token and POS tag uigram, POS tag bigram, phrase pattern) and classifiers (RF, SVM, LMT, ADD, DWD, HDD). This approach was tested on three text types: novels, student essays, and personal diaries. Four types of features and six classifiers were used to classify the texts. An ensemble of the 24 results derived from four types of features and six classifiers demonstrated excellent validity.

The scores of ensembles of multiple classification models generally equal or exceed the best score achieved by any individual model; moreover, ensembles are more robust than individual models. Robustness is crucial for drawing conclusions to real-world problems. Hence, ensembles are gradually finding practical applications [17,18]. Bacciu et al. [17] performed text classification in a soft-voting ensemble using the classification results of multiple different features with a single classifier (SVM). Lasotte et al. [18] used the results of four classifiers with a single feature dataset to detect fake news and demonstrated the effectiveness of ensembles. 

In 2013, Mikolov et al. [19] proposed an algorithm for embedding words or sentences into vectors, demonstrating an approach for classifying text without extracting its features. In 2018, Google released a BERT model pre-trained on a large corpus of English text (Wikipedia + BookCorpus); the model achieved state-of-the-art (SOTA) performance on several natural language processing (NLP) tasks [20]. 

BERT comprises a transformer architecture with a deep neural network structure, which is trained using large-scale data. It embeds words and their contextual relations by quantifying them into fixed-length vectors. The BERT model is divided into a base model and a large model, depending on the layers of the neural network used for deep learning. The BERT-base model has 12 layers of encoders (transformer blocks) and 768 hidden layers. Conversely, the BERT-large model has 24 layers of encoders and 1024 hidden layers. BERT-large has more parameters and can be adapted to more complex NLP tasks. However, it requires more computational resources and time. 

The BERT model depends on the corpus and language type used for pre-training. End users fine-tune the models with the corpus of their task and then adapt BERT. Since the study by Devlin et al. [20], various models have been proposed globally and validated for a wide range of tasks, including question–answer classification, classification of newspaper news types, sentiment analysis, finance, and clinical and medical tasks. Although many classifiers and PLMs have been developed, different aspects have been studied regarding the best model for a given task.
	
\subsection{Comparison of PLMs and classifiers}
Numerous pre-trained BERT models have been introduced, and empirical studies have been conducted to determine the best BERT model for specific tasks. Tyo et al. [21] evaluated eight promising classification models using 15 datasets. The results indicated that the method using conventional n-gram-based features was better than BERT for the AA task on datasets with small total samples per author, whereas the BERT-based model was superior on the datasets with large samples. The smallest dataset contained 5000 texts. 

Qasim et al. [22] conducted a binary classification of fake news using two datasets (COVID-19 fake news and extremist–non-extremist datasets) with nine different BERTs. The sample size was over 10,000. BERT and RoBERTa scored differently depending on the dataset, and no conclusion could be drawn as to which one was better. Similarly, the scores of base and large models varied, depending on the dataset, and the scores of large models were not always higher than those of base models.

 Karl and Scherp [23] performed a comparative analysis of 14 (BERT-, BoW-, graph-, and LSTM-based) models. The SOTA performance depended on the dataset. However, even the smallest of the four datasets exceeded 8000. 

Sun et al. [24] and Zhang et al. [25] used six and eight PLMs, respectively. In the benchmark data used, the SOTA performance of the BERT models was dataset-dependent. Nevertheless, the mean accuracy of DeBERTa was slightly better than that of RoBERTa [25]. The smallest sample size among the seven datasets used was over 7000. 

Prytula [26] compared and analyzed the binary classification of positive and negative user comments on a dataset of approximately 11,000 user comments written in Ukrainian, comparing the BERT, DistiIBERT, XLM-RoBERTa, and Ukr-RoBERTa models. The results indicated that the XLM-RoBERTa model achieved the highest accuracy. However, considering the time required to train the model and all the classification indices, Ukr-RoBERTa was the best. 

Abbasi et al [41] compared the performance of DistilBERT, BERT, and an ensemble of classifiers using features to determine the authors of news articles. Their results showed that DistilBERT outperformed both BERT and the feature-based ensemble. They used 50,000 articles to classify 10 or 20 authors.
	
\subsection{Effect of pre-training data on the BERT model}
Mishev et al. [27] conducted a comparative analysis of 29 models including feature-based methods and BERT for sentiment analysis in finance using two datasets: Financial Phrase-Bank and SemEval-2017 Task5. The results indicated that FinBERT pre-trained on Reuters financial datasets did not score as well as BERT pre-trained on Wikipedia and BookCorpus.

Arslan et al. [28] used BBC News and 20News datasets to compare the performance of FinBERT built from financial datasets and five general-purpose PLMs (BERT, DistilBERT, RoBERTa, XLM, and XLNet) built from other pre-trained data such as Wikipedia. They found that RoBERTa, built from a corpus of nonfinancial corpora, was superior to other models. More than 2000 data samples were used, even for small corpora.

Suzuki et al. [29] created a FinDeBERTaV2 model using a financial dataset and conducted a comparative analysis with GenDeBERTaV2. The results demonstrated that GenDeBERTaV2 performed better on general problems, whereas FinDeBERTaV2 performed better in the financial domain.

Ling [30] compared the effectiveness of PLMs, including BERT-base and Bio+Clinical BERT, in classifying the treatment sentiments of drug reviewers. The data were obtained from the UCI ML Drug Review dataset of 215,063 drug reviews. The Bio+Clinical BERT classification score on the test data surpassed that of the BERT base model.

Vanetik et al. [31] classified the genre of Russian literature using stylistic features and BERT. The results indicated that the scores of the classifier with stylistic features were higher than those of BERT. Experimental results indicate that ruBERT, which was pre-trained on a large Russian corpus, performed more poorly than the multilingual BERT model. The SONATA dataset used was a random sample of 11 genres with a sample size of 100 for each genre, and the chunks were extracted manually. 

Thus, the findings on the influence of pre-trained corpora on specific tasks are conflicting. This remains an area that requires further investigation. Kanda and Jin [32] conducted a comparative analysis of the effects of pre-training data on tasks using literary works for four different types of BERT, including BERT using literary works, and showed in their interim report in Japanese that pre-training data had a significant effect on tasks. Additional experiments and analysis are presented in Subsection 4.1.
	
\subsection{Ensemble of BERTs}
An ensemble of BERT models was also discussed in Devlin’s [20] study, which originally introduced BERT. Ensembling BERTs involves (1) ensembling the results of different models obtained by changing checksheets or datasets when fine-tuning, and (2) ensembling the results of multiple BERT models created with different pre-training data and parameters. 

Tanaka et al. [33] conducted a study using BERT for Japanese texts longer than 510 tokens. They first cut 510 tokens while shifting them at regular intervals toward the end of the text to create several different samples. Subsequently, they ensembled the BERT results obtained on these different samples. The ensemble scored better than the single BERT using only 510 tokens. In their experiment, they used three datasets, each with 2000 samples, for training and testing.

Xu et al. [34] ensembled multiple BERTs trained on a question–answering task using different dataset sizes and batch files. They reported that ensemble results yielded the highest scores. The dataset they used had more than 100,000 samples. Dang et al. [35] won the SMM4H Task1 competition with an ensemble of 20 results obtained with 10 sets of samples from 10 cross-validations and two different BERTs.

Abburi et al. [36] won the first prize in the 2023 competition for their study on the identification of LLM-generated texts using an ensemble of DeBERTa, XLM-RoBERTa, RoBERTa, and BERT. The ensemble involved using a classifier to classify a vector of the BERT results. The data used were from four datasets distributed in the competition, with the smallest sample size exceeding 20,000.

Zamir et al. [37] conducted an empirical study on the optimization of ensemble learning using BERT, ALBERT, DistilBERT, RoBERTa-base, and XLM-RoBERTa, and achieved improved performance when using the optimized ensemble. The reported results showed no significant difference between the optimization methods. The data used were the dataset of the author analysis shared task distributed in the PAN-21 competition, with a sample size of over 10,000.

\subsection{Ensemble of BERT and features}
Tanaka et al. [33] demonstrated that text classification using a three-layer neural network on data consisting of BoW vectors concatenated to BERT-embedding vectors improved the classification scores. For 510 tokens, the score increased by 2.1 points.

 Fabien et al. [38] used logistic regression on the classification results of a single BERT, stylistic features, and features related to sentence structure. Consequently, most of the ensemble scores were lower than the scores of the single BERT. Four datasets were used in the experiment, and the number of training samples per author was not more than 100 in a few cases. 

Wu et al. [39] won the competition by ensembling the results of BERT and RoBERTa on a disease-related question-and-answer dataset with those of the tree model on a single dataset extracted from text. The ensemble's F1 score was 3.38 points higher than that of the best-performing model, RoBERTa\_wwm. The sample size of the training data used was 30,000. 

Strøm [40] detected author style changes by stacking ensembles of a set of 956 dimensional stylistic features and 768 dimensional embedding vectors by BERT; they obtained the best score in the competition for the classification task in the same year. However, the scores increased by only 1.93, 2.86, and 1.01 points, compared with those of the LightGBM classifier for the three tasks. The sample size of the dataset was over 10,000.

However, the following issues arise when the aforementioned studies are considered from the perspective of authorship analysis. \\
\textbf{(1) Sample issue:} The aforementioned BERT-related research reports show that the data samples used in the evaluation or application of BERT models are typically in the order of a thousand. In Devlin’s study [20], the size of the smallest dataset in the GLUE benchmark was 2.5K. Although larger samples are common when building large-scale classification systems, real-world problems, such as those related criminal or civil law in forensic science, often involve far fewer samples, sometimes fewer than 10. This highlights the need for further research on small samples. However, no studies have addressed this challenge. \\
\textbf{(2) BERT and the issue of pre-trained data:} Although many pre-trained models have been proposed, no conclusion has been drawn regarding the best model. This is because the performance of the models depends on the sample size used in the test, type of task, and parameters specified during tuning. Therefore, for end users, continued research should be conducted on the types of models to be used and how to effectively utilize existing data when applying BERT. Several studies have been conducted on the effect of pre-training data on tasks; however, no clear conclusions have been drawn. Furthermore, there have been no studies on literary works. \\
\textbf{(3) Ensemble Issue:} We did not find any studies in which the results of multiple BERTs were trained on different pre-training data using the results of multiple feature datasets and classifiers. In most of the above studies, the features used were concatenated into a single vector. Unlike topic classification, several stylistic features have been proposed for AA; however, it is difficult to determine the best. This is because stylistic features depend on the writer’s genre and style. In addition, although it is easy to collect high-dimensional data, concatenating them into a single dataset significantly affects the classifier performance. For example, aggregating bigrams of characters yields thousands of dimensions. It is not recommended to use ultra-high-dimensional data created by combining multiple feature datasets with thousands of different dimensions side by side when working with small-scale samples of only tens or hundreds of units. In addition, studies have explored how to use the many variations of BERT models and feature-based models available. However, they offer only marginal and unimpactful score improvements. 

Considering these issues, we focused on significantly improving author estimation scores in the AA tasks on small samples and literary works using an integrated ensemble of BERT- and feature-based results. We also analyzed the effect of pre-training data on the task and the effect of the model characteristics used for the integrated ensemble.

\section{Experiment}
Based on previous research, we classified two corpora using multiple BERT models, features, and classifiers; then, we ensemble-integrated the classification results. The workflow of this study experiment is shown in Fig 1. 

\begin{figure}[h]
  \centering
  \includegraphics[width=0.4\textwidth]{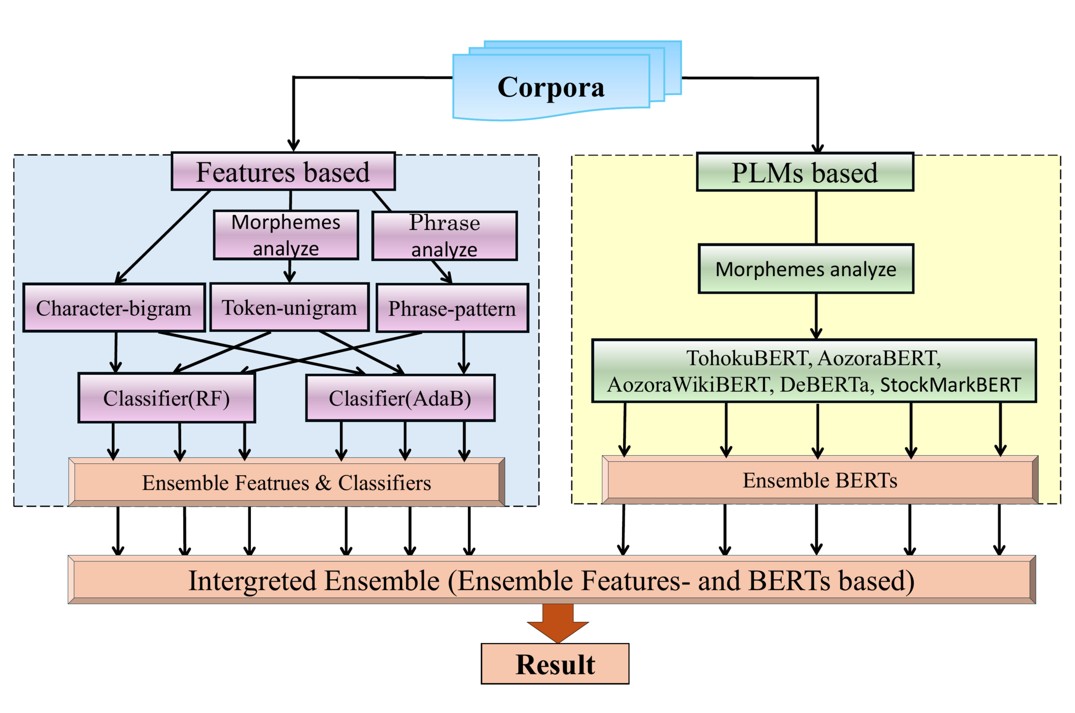} 
  \caption{The overview of workflow of this study}
  \label{fig:example}
\end{figure}

The basic concept underlying these ensembles is collective intelligence. Since the 1990s, ensemble learning has been used in the field of machine learning to improve performance by combining the results of multiple models. Ensembles include voting and stacking methods; however, voting methods are most often used when integrating the predictions of the base model. There are two types of voting: hard and soft. Hard voting involves aggregating the prediction results of each model, then selecting the class with the most votes as the final prediction. This is also called majority voting. Soft voting involves summing or weighting the probability scores output by each model, then using the score with the highest probability value as the final forecast result.

 In this study, we used an ensemble method based on soft voting. Specifically, the probability vectors, $Model^k(x)$, obtained from different models (classifiers or BERT) are summed or averaged as shown in Equation (1), and the author with the largest value is assigned as the author of index $j*$. 
In the equation, $w_k$ is the weight of classification model $k$. We assigned a vector of 1 when we did not consider the weights, and used the F1 score of each model when weights were considered.

\begin{equation}
 j^*=\underset{j}{argmax} {[\frac{1}{K} \sum_{k=1}^K w_k Model^{k}(x)]}
\end{equation}

In this study, the results of individual classifiers were probability vectors, and the results of each BERT were converted into probability vectors using a Softmax function.

\subsection{Features and classifiers used}
Several features have been proposed for the quantitative analysis of writing style and author estimation. We used features on those demonstrated by [10, 11, 44]. We used character-bigram, token-unigram, and phrase pattern (Bunsetsu-pattern) as representative stylistic features.\\
\textbf{(1) Character-bigram}\\
Character n-gram is a dataset that aggregates the patterns of n adjacent characters. In this study, to estimate authorship using short sentences of 510 tokens (approximately 800 characters), we used the most frequently used characters bigram [15, 42, 43] based on the dimensionality and sparsity of the data. \\
\textbf{(2) Token-unigram}\\
Morphemes were analyzed using the MeCab [R1] and UniDic [R2] dictionaries. Among the n-grams of token, we used the unigram, which mainly reflects the characteristics of an author [14, 15].  \\
\textbf{(3) Phrase-pattern}\\
Many indicators of authorship are also found in the sentence syntax. The basic unit of Japanese parsing is a phrase (Bunsetsu). Jin [44] proposed and validated a method that patterned information in a phrase. 

 "\begin{CJK}{UTF8}{ipxm}To illustrate for Japanese, consider a sentence,"BERTは分類に有効である。" (BERT is effective for classification). The sentence is split into phrases as "BERT (noun)は (particle)/分類 (noun)に (particle)/有効 (adjectivalNoun)で (adjective)ある(verb)。 (punctuation)". (BERT is/ effective for/ classification). In this context, the symbol "/" indicates the phrase boundaries, and the string within the parentheses represents the POS tag of the preceding morpheme. 

The first phrase in the sentence, "BERT (noun)は (particle)," consists of a noun "BERT" and a particle "は". The word "BERT" is content-dependent; therefore, if it is used as a feature in the AA, it will be a noise in the analysis of the author's characteristics. Therefore, the POS tag noun is used in BERT to mask the content word. The result, "BERT (noun)は (particle)," is patterned after "noun + は(particle)". The particle "は" may change depending on the author. This is because using "が"(ga) instead of "は"(wa) does not pose a grammatical problem. Words that are not characteristic of the writer are masked with their POS tags in phrase patterns. The data related to phrase patterns is an aggregation of such patterns categorized by type.
\end{CJK}

 In Japanese, the usage data of particles and punctuation have been demonstrated to be effective features for AA in numerous studies [2]. The phrase pattern can capture individual habits, specifically how distinctive elements of a writer's style are combined and used. Although a phrase pattern never scores high in author estimation, it is robust to text content and genre because content words are masked by their POS tags [10, 11, 15]. In a study on AA for texts containing mixed genres [15], the combination of phrase pattern features and a Lasso classifier achieved the highest accuracy (0.895) among 14 feature types and seven classifiers tested in the comparative analysis.

In this experiment, CaboCha [R3] was used for the phrase segmentation process. Morphemes were masked using POS tags—excluding particles, punctuation marks, conjunctions, and adjectives—based on the syntactic parsing results generated by CaboCha. \\
\textbf{(4) Classifiers}\\
We used two types of classifiers, Random Forest (RF) and AdaBoost (Ada), which have been reported to achieve relatively high performance. For RF, we used the randomForest package in R, an algorithm proposed by Breiman [45]. However, although XGBoost is reported to have excellent performance among boosting algorithms, our study, which focused on using stylistic features, did not obtain any results that demonstrated XGBoost’s superior performance over AdaBoost [15, 46]. In this study, we used the adabag package in R for AdaBoost [47]. The default parameters were used for both algorithms.

\subsection{Selected PLMs}
In this study, we used several BERT models as PLMs. BERT is a pre-trained language model built on the Transformer architecture, taking a fundamentally different approach from traditional classifiers. Traditional classifiers rely on extracting features from text, representing them as vectors, and applying models like logistic regression. These methods focus solely on feature extraction based on frequency, ignoring sequential and contextual information.

In contrast, BERT generates contextualized vector representations through a multi-layer Transformer network pre-trained on large-scale text corpora. These embeddings encapsulate both semantic meanings and positional information as fixed-length vectors (e.g., 768 dimensions), capturing context-dependent word relationships learned during the pre-training phase. A key innovation of BERT lies in its self-attention mechanism, which bidirectionally analyzes all token pairs within a sentence. By calculating attention weights that reflect inter-token relevance, the model prioritizes semantically critical connections in a context-dependent manner. This design enables robust context modeling, effectively handling even long-range dependencies that challenge traditional sequential models.

BERT's pre-training involves two core tasks: Masked Language Modeling (MLM) and Next Sentence Prediction (NSP). MLM randomly masks tokens and trains the model to predict the masked positions based on bidirectional context. In contrast, NSP learns inter-sentence coherence by predicting whether two sentences are consecutiv. Through these tasks, BERT acquires deep linguistic understanding from unlabeled text data.

The pre-trained model can then be fine-tuned for downstream tasks such as text classification. BERT classifies text by embedding vectors, which represent word meanings and context, into a neural network to predict class labels. This transfer learning paradigm achieves high accuracy even with limited labeled data, overcoming the data scarcity limitations of traditional approaches.

We used BERTs built on Japanese pre-training data, which are described below. The multilingual XLM-RoBERTa has been released as an advanced variant of BERT. XLM-RoBERTa uses SentencePiece, a tokenizer designed to maintain consistency across different languages, thus not fully leveraging the characteristics of the Japanese language. Additionally, the accuracy of SentencePiece has been reported to be lower than that of a Japanese-specific tokenizer, MeCab. Therefore, this study used a BERT model specialized for the Japanese language.

The following five BERT models were selected based on the diversity of the pre-training data used for training and their performance. In addition, we are limited to BERTs trained using pre-training data processed with tokenizers that are highly rated against the Japanese language.\\
\textbf{(1) Japanese BERT trained on Wikipedia}\\
There are many BERTs pre-trained on the Japanese Wikipedia. Based on preliminary analysis, we used the most widely used basic version published by Tohoku University that base version of a publicly available model pre-trained on data tokenized using MeCab and WordPiece. 

There are many BERT models pre-trained on the Japanese Wikipedia. Based on preliminary analysis, we adopted the most widely used basic version — published by Tohoku University as the foundation of a publicly available model — which was pre-trained on data tokenized with both MeCab and WordPiece. Implementation details are provided in [R4].\\
\textbf{(2) Japanese BERT trained on Aozora Bunko}\\
Koichi Yasuoka has released a pre-trained BERT model trained on the Aozora Bunko corpus — a public-domain repository of Japanese literary works. From the published variants, we adopted the model pre-trained using MeCab tokenization with the UniDic dictionary. Implementation details are provided in [R5]\\
\textbf{(3) Japanese BERT trained on Aozora Bunko and Wikipedia}\\
Koichi Yasuoka has released BERT models pre-trained on a combined corpus of Aozora Bunko and Wikipedia.  From these, we adopted a model pre-trained on data tokenized using MeCab with the UniDic dictionary. Implementation details are provided in [R5].\\
\textbf{(4) DeBERTa}\\
He et al. [48] proposed DeBERTa, an improved version of the BERT and RoBERTa models.　 DeBERTa introduces Disentangled Attention, a mechanism that separately encodes word content and positional information into distinct vectors. This separation enhances contextual representation learning, leading to superior performance over standard BERT architectures. Furthermore, DeBERTa employs an Enhanced Mask Decoder during pre-training, which explicitly incorporates absolute positional data to better align contextual and positional features. The objective of this design is to improve the prediction accuracy of masked tokens compared to other conventional BERT models.

We used the Japanese DeBERTa V2-base model pre-trained on Wikipedia, CC-100, and OSCAR corpora with JUMAN++ tokenization. Implementation details are provided [R6]. \\
\textbf{(5) BERT StockMark} \\
StockMark Inc. released a BERT model pre-trained on Japanese news articles. As new words are created annually in business news, unknown words are processed as [UNK] tokens without subword settings. We used a model pre-trained on data tokenized using MeCab with the NEologd dictionary. Implementation details are provided in [R7].

For brevity, we used the abbreviations TohokuBERT (T), AozoraBERT (A), AozoraWikiBERT (AW), DeBERTa (De), and StockMarkBERT (S) for the BERTs.

\subsection{Corpora used}
Two corpora were used in this study: Corpora A and B. Corpus A consisted of 10 authors selected from the literary authors available in the Aozora Bunko, and 20 works for each author. The amount of Corpus A data is approximately 0.03\% of the pre-training data in Aozora Bunko. Specifically, the corpus contained 20 works by Akutagawa Ryunosuke (1892–1927), Izumi Kyoka (1873–1939), Kikuchi Kan (1888–1948), Mori Ogai (1862–1922), Natsume Soseki (1867–1916), Sasaki Ajitsuzo (1896–1934), Shimazaki Toson (1872–1943), Dazai Osamu (1909–1948), Kido (1872–1939), and Juzo (1897–1949). The works were prioritized by those that had already been converted to the new script and new kana usage and those that were published in the same year.

Corpus B was an electronic version of 20 works in paper form by 10 writers who are still active in the field and were not used for the pre-training of BERT. Specifically, the authors included Kouji Suzuki (1957-), Yusuke Kishi (1959-), Shuichi Yoshida (1968-), Miyabe Miyuki (1960-), Morimi Tomihiko (1979-), Ishida Ira (1960-), Murakami Haruki (1949-), Murakami Ryu (1976-), Higashino Keigo (1958-) and Minato Kanae (1973-).

We conducted experiments to attribute 10 authors each using these two corpora of literary works.

\subsection{Experimental setup and metrics for evaluating results}
Generally, $k$-fold cross-validation was used to evaluate the classification results by the supervisor. Following previous research, we used the five-fold cross-validation. Typically, $k$-fold cross-validation involves randomly dividing the data into k subsets. However, owing to the small sample size of our data, random division may bias the balance of class labels. Therefore, in this experiment, we used a stratified sampling method to divide the dataset into five folds in the following proportions: training data (160), validation data (20), and test data (20) per fold. The number of works by each author for learning and testing in all the folders was designed to ensure balance.

In languages where texts are not inherently segmented into tokens, such as Japanese and Chinese, processing with BERT necessitates tokenization at either the character or morpheme level. Although character-level BERT models for Japanese demonstrate suboptimal performance, current implementations preprocess texts via morphological analysis, treating each morpheme as a discrete token. 

In this study, we morphologically analyzed 200 works each from Corpus A and Corpus B using MeCab for morphological segmentation. Each morpheme was treated as a single token, with the first 510 tokens extracted and formatted as input text for BERT. To meet BERT's maximum input length of 512 tokens, [CLS] and [SEP] tokens were prepended and appended, respectively, during the conversion of tokenized sequences into input IDs. BERT then converted the input into vector embeddings through deep learning and output classification results based on those vectors. In this study, for fairness, only the first 510 tokens were used for all processes, including feature-based methods.

Based on a preliminary analysis of the hyperparameters for fine-tuning BERT, we set the mini-batch size to 16 and learning rate to 2e-05 for any BERT and used AdamW as the optimization algorithm. For the epoch, we set the value to 40 after maintaining a stable value for all the BERTs because of the relatively small number of data samples used. 

The most widely used performance measure of a model is the macro average of the F1 measure, which balances the recall and precision measures. The equations for each evaluation index are as follows. The estimation results for author $i$ ($i = 1, 2, 3,  \ldots, M$) and other authors are shown in the confusion table in Table 1.

\begin{table}[h]
\centering
\caption{Confusion table of classification results}
\vspace{0.2cm}
\label{tab:confusion_matrix}
\begin{tabular}{|c|c|c|}
\hline
For Author $i$ & Pred. Po.& Pred. Neg.\\ \hline
True Positive & $TP_i$ & $FN_i$ \\ \hline
True Negative & $FP_i$ & $TN_i$ \\ \hline
\end{tabular}

\end{table}
\vspace{-0.6cm}

\begin{equation} 
\text{Recall}_i = \frac{TP_i}{TP_i + FN_i} 
\end{equation}
\vspace{-0.6cm}

\begin{equation}
\text{Precision}_i = \frac{TP_i}{TP_i + FP_i}
\end{equation}
\vspace{-0.6cm}

\begin{equation}
\text{F1}_i = 2 \times \frac{\text{Precision}_i \times \text{Recall}_i}{\text{Precision}_i + \text{Recall}_i}
\end{equation}
\vspace{-0.2cm}
\begin{equation}
\text{Macro F1} = \frac{1}{M} \sum_{i=1}^{M} \text{F1}_i
\end{equation}

The $\text{Recall}_i$, $\text{Precision}_i$, and $\text{Macro F1}_i$ mentioned above were the metrics used to evaluate the performance of the classification model results for each individual class (author) $i$. $M$ is the number of classes (authors).

$\text{Recall}_i$ (True Positive Rate) is the proportion of actual positives for class i correctly identified by the model. $\text{Precision}_i$ is the proportion of predicted positives for class i that are true positives. $\text{F1}_i$ is the harmonic mean of $\text{Precision}_i$ and $\text{Recal}_i$, providing a balanced measure of the model's performance for class $i$. Hereafter, Macro F1 is abbreviated as F1.

\section{Results and analysis}
\subsection{BERT results}
We evaluated the performance of the BERT models on the test data at the point when the performance stopped improving on the validation data. Table 2 presents the experimental results of BERT on the corpora. The highest F1 scores for both corpora are shown in bold.

 For Corpus A, Model AW pretrained on Wikipedia and Aozora Bunko had the highest score and smallest standard deviation, followed by Model A pre-trained on Aozora Bunko. Both models achieved F1 scores of more than 30 points, which was singularly higher than that of Model T, which used only Wikipedia. It is natural to assume that this is because Corpus A was included in the pre-trained data. 

In Corpus B, Model De had the highest F1 score, followed by AW and A. The model with the lowest score in both corpora was S, which was pre-trained using business news articles. The fact that Model De achieved the highest score for Corpus B indicates that Corpus B was not included in the pre-training data, and Model De was more generic. The scores of Models AW and A were higher than those of the models pre-trained using Wikipedia and news articles because Corpus B is also a literary work. We believe that this performance difference is owing to the influence of text style. 

The scores of Model S pre-trained using Wikipedia and Japanese business news articles were higher for Corpus B than for Corpus A. This variation could be attributed to the differences in vocabulary and grammar used in the period, as most of the works in Corpus B were published in the 1990s. These results suggest that pre-trained data affect individual tasks. Generally, BERT performs better when the data size is large and data from many domains are used for pre-training; a similar trend was observed in the results of this study.

\begin{table}[h]
\centering
\caption{Results of 10 author discrimination \\
by BERT}
\vspace{0.2cm}
\label{tab:confusion_matrix}
\small
\begin{tabular}{ccccc}
\hline
Corpurs & BERT & Recall & Precision & F1 \\ \hline
 					& T  & 0.653  & 0.640  & 0.642 \\ 
 					& A  & 0.973  & 0.970  & 0.969 \\ 
A				& AW & 0.973  & 0.970  & \textbf {0.970}\\ 
   				&De & 0.752  & 0.680  & 0.691\\ 
			 		&S   & 0.619  & 0.600  & 0.600  \\ \hline
			 		& T    & 0.762  & 0.740  & 0.744 \\ 
 					& A    & 0.813  & 0.770  & 0.773  \\ 
B				& AW & 0.838  & 0.820  & 0.820  \\ 
					& De  & 0.834  & 0.820  & \textbf {0.823} \\ 
 					& S    & 0.706  & 0.690  & 0.692  \\ \hline
\end{tabular}
\end{table}

\subsection{Features and classifier results}
Experiments on feature extraction from literary works and identification of authors using classifiers were conducted under the same conditions, including the length of the works used, as those in the BERT experiment. The dimensions of the extracted feature datasets were 4444 for the char-bigram, 3300 for token-unigram, and 804 for phrase pattern. The classification results of the Ada and RF classifiers for these feature datasets are listed in Table 3. The extraction of features from the text and processing using R was done using MTMineR [50, R8]. The average scores of both corpora did not differ significantly. The highest F1 score was obtained by the RF using token-unigram, and the lowest score was obtained by the RF using phrase pattern. 

\begin{table}[h]
\centering
\caption{Results of 10 authors discrimination\\
 by features and classifiers}
\vspace{0.2cm}
\label{tab:classification_results}

\small 
\begin{tabular}{ccccc}
\hline
Corpurs   & Clf. and Feat. & Recall & Precision & F1 \\ \hline
    			  	& Ada-Char 		  & 0.786  & 0.760       & 0.766\\
				    	& Ada-Token    & 0.767& 0.750 & 0.754 \\ 
 	A  		  	& Ada-Phrase   & 0.762& 0.750 & 0.747 \\ 
					    & RF-Char		    & 0.792 & 0.790 & 0.784 \\ 
 					    & RF-Token     & 0.823 & 0.810 & \textbf {0.810} \\ 
 					    & RF-Phrase    & 0.714  & 0.710 & 0.704  \\ \hline
			        & Ada-Char  	  &	0.779 & 0.760 & 0.761  \\ 
				      & Ada-Token    & 0.772 & 0.760 & 0.762  \\ 
 	B			    & Ada-Phrase& 0.654 & 0.650 & 0.647  \\ 
	  		     	& RF-Char     & 0.780 & 0.780  & 0.767  \\ 
 				      & RF-Token   & 0.810  & 0.800 & \textbf {0.800} \\ 
 				      & RF-Phrase  & 0.668 & 0.650& 0.643  \\ \hline
\end{tabular}
\end{table}

\subsection{Ensemble of BERTs}
There were 26 ways to arbitrarily ensemble two or more of the five BERTs. For the weighted ensemble, we used the F1 scores for each BERT model. To save space, the summary statistics of the F1 scores for the ensembles of both methods are presented in the second row of the corpus in Table 4.

 For Corpus A, the maximum score of the ensemble and the mean increased by two points and 13.7 points, respectively. For Corpus B, the maximum score of the ensemble and the mean increased by 7.9 and 9.2 points, respectively. The weighted ensembles did not show any increase in score compared to the unweighted ensembles in either corpus. 

Direct comparison with existing AA methods is challenging owing to their reliance on large, publicly available datasets, primarily in English. To facilitate a meaningful comparison, we adapted established, reproducible methods to our corpus. The results are presented in Table 4.

To consider the combinatorial situation in the ensemble, the F1 scores of the top 10 sets of ensembles are listed on the left side of Table 5. For Corpus A, \{A, S\}, \{T, A\}, \{A, AW\}, \{A, De\}, \{A, AW, De\}, and \{AW, S\} exceeded the maximum F1 score of a single BERT. In Corpus B, 22 ensembles exceeded the maximum value of 0.820 for the BERT model. The highest scores were obtained for \{T, A, AW, De\}, followed by \{A, AW, De, S\}, \{T, A, De, S\}, and \{T, A, AW, De, S\}. 

Interestingly, the combination with the highest ensemble F1 score in both corpora included Model S, despite it having the lowest individual score among all models. News articles that are vastly different from the novel-writing style were used for pre-training Model S, which is the subject of this task. This suggests that although ranking individual scores is important when performing ensembles, it is also important to combine heterogeneous models. 

For both corpora, the ensemble scores of \{T, A\} were 0.98 and 0.856, respectively, which are higher than those of Model AW, i.e., 0.97 and 0.82, respectively. Wikipedia and Aozora Bunko were used for pre-training T and A, respectively, and Wikipedia and Aozora Bunko for Model AW. However, the ensemble scores of Models T and A were higher than those of Model AW. This again suggests that the performance of the models is affected by the data used for pre-training and other properties of the models.

Table 4. Statistics of F1 values for BERT-based and feature-based ensembles and integrated ensembles (The mean and standard deviation(sd) of the top.

\begin{table*}[h]
\caption{Statistics of F1 values for BERT-based and feature-based ensembles and integrative ensembles (The mean and standard deviation(sd) of the top 50 are used for methods with F1 values greater than 50.)}
\label{tab:statistics}
\vspace{0.3cm}
\centering
\small 
\begin{tabular}{ccccc}
\midrule
        													 &  \multicolumn{2}{c}{Corpus A}& \multicolumn{2}{c}{ Corpus B} \\ 
Method													 &   Mean ± sd		& Max        & Mean ± sd   & Max              \\ \midrule
BERTs & 0.775 ± 0.181 & 0.970 & 0.770 ± 0.055 & 0.823 \\ 
Ensemble BERTs [32, 36] & 0.911 ± 0.091 & 0.990 & 0.861 ± 0.030 & 0.902 \\ 
Weighted Ensemble BERTs [37] & 0.910 ± 0.096 & 0.980 & 0.861 ± 0.029 & 0.899 \\ 
Features \& Classifiers & 0.761 ± 0.036 & 0.810 & 0.730 ± 0.067 & 0.800 \\ 
Ensemble Features \& Classifiers [16-18, 41] & 0.852 ± 0.033 & 0.912 & 0.817 ± 0.039 & 0.889 \\ 
Weighted Ensemble of Features \& Classifiers & 0.851 ± 0.034 & 0.912 & 0.828 ± 0.033 & 0.889 \\ 
Ensemble One Feature \& Classifiers and BERTs [39] & 0.934 ± 0.040 & 0.970 &0.887 ± 0.044 & 0.920\\ 
Ensemble One BERT and Features \& Classifiers [40] & 0.834 ± 0.127 & 0.990 & 0.814 ± 0.052 & 0.901 \\ 
Integrated Ensemble & 0.991 ± 0.003 & \textbf {1.000} & 0.957 ± 0.005 & \textbf {0.960} \\ 
Integrated Weighted Ensemble & 1.000 ± 0.000 & \textbf {1.000} & 0.953 ± 0.005 & \textbf {0.960} \\ \midrule
\end{tabular}
\end{table*}

\subsection{Ensemble results for features and classifiers}
The ensemble of the six results (1: Ada+Char, 2: Ada+Token, 3: Ada+Phrase, 4: RF+Char, 5: RF+Token, and 6: RF+Phrase) of the two classifiers (Ada and RF) with three features (char-bigram, token-unigram, and phrase pattern) yielded 57 results.

The summary statistics of the ensemble results for both corpora are presented in Table 4. For both corpora, the maximum F1 scores of the ensembles were significantly higher than those of the stand-alone features and classifiers. For Corpus A, the maximum F1 score was 10.1 points higher than those of the single features and classifiers, and the average score was 9.1 points higher. For Corpus B, the maximum F1 score was 8.9 points higher than those of the single features and classifiers, and the average score was 8.7 points higher. Table 5 shows the top 10 scoring ensembles. Some combinations of features and classifiers in the ensemble included Labels 3(Ada+Phrase) and 6(RF+Phrase), which had the lowest scores. The reasons for this will be analyzed considering the results of the integrated ensemble.

\subsection{Integrated ensemble}
The ensemble results from BERTs and those from the features and classifiers were integrated-ensembled. There were 26 ensemble results for the five BERT models and 57 ensemble results for the three features and two classifiers. Ensembling these results further yielded 1482 ensembles. “Integrated ensemble” refers to the ensemble of all results obtained from various aspects. 

The statistics of the F1 scores for the integrated ensembles are presented in Table 4, where the results for the integrated ensemble are the top 50 statistics. 

For Corpus A, the highest F1 score was 1. For Corpus B, the highest F1 score was 0.96. For Corpus A, the F1 score was 19 points higher than that of the single model with features; for Corpus B, it was 13.7 points higher than the maximum score of the single model, confirming the effectiveness of the integrated ensemble. In addition, the integrated-ensemble F1 scores in both corpora improved by one and two points, respectively, compared to the ensemble of the results of the BERTs used and a single feature and classifier, or the ensemble of the result of a single BERT and that of the features and classifiers used. The results for the weighted ensemble were almost the same. 

For comparison, results of both ensemble methods in [39, 40] were computed and summarized in Table 4. The scores significantly improved over single models, BERT ensembles, and feature-based ensembles in both corpora, but they did not reach the results of the proposed integrated ensemble.

Fig 2 presents the box plots of F1 scores for both corpora. The top 50 results are shown for methods with F1 scores greater than 50. The integrated ensemble(I) demonstrates significantly higher F1 scores and a substantially smaller score variance compared to baseline methods, as evidenced by the box plot distributions.

\begin{figure}[h]
  \centering
  \includegraphics[width=0.5\textwidth]{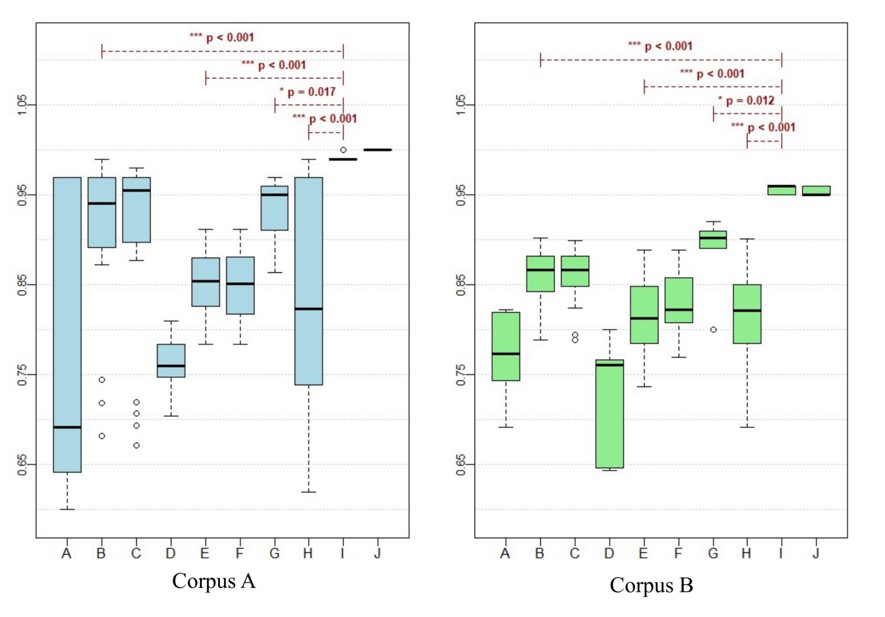} 
  \caption{Box plot of F1 scores for both corpora. The correspondence between the horizontal axis labels and datasets is as follows: A: BERTs, B: Ensemble BERTs[32, 36], C: Weighted Ensemble BERTs, D: Features \& Classifiers, E: Ensemble Features \& Classifiers[16-18, 41], F: Weighted Ensemble Features \& Classifiers, G: Ensemble One Feature \& Classifiers and BERTs [39], H: Ensemble One BERT and Features \& Classifiers [40], I: Integrated Ensemble, J: Integrated Weighted Ensemble.}
  \label{fig:example}
\end{figure}

To rigorously quantify the significance of improvements specifically for ensemble-based approaches, we conducted Welch's two-sample t-tests to compare the Integrated Ensemble (I) against four baselines: Ensemble BERTs[32, 36] (B), Ensemble Features \& Classifiers[16-18, 41] (E), Ensemble One \& Classifiers with BERTs [39] (G), and Ensemble One BERT and Features \& Classifiers [40] (H). The results of Welch's two-sample t-tests for Corpus A are as follows: 

\setlength{\itemsep}{0.1em} 
\setlength{\parsep}{0pt} 
\begin{itemize}[itemsep=0.1em, parsep=0pt]
    \item I vs. B ($p=0.0001$, Cohen's $d=0.880$)
    \item I vs. E ($p<2.2\times10^{-16}$, Cohen's $d=4.168$)
    \item I vs. G ($p=0.017$, Cohen's $d=4.546$)
    \item I vs. H ($p=3.2\times10^{-7}$, Cohen's $d=1.202$)
\end{itemize}

For Corpus B are as follows: 

\begin{itemize}[itemsep=0.1em, parsep=0pt]
    \item I vs. B ($p=4.2\times10^{-15}$, Cohen's $d=3.232$)
    \item I vs. E ($p<2.2\times10^{-16}$, Cohen's $d=3.631$)
    \item I vs. G ($p=0.012$, Cohen's $d=4.939$)
    \item I vs. H ($p=3.3\times10^{-15}$, Cohen's $d=2.718$)
\end{itemize}

All pairwise comparisons except I vs. G (both corpora) showed statistically significant differences ($p < 0.001$). For the I vs. G comparison, the p-value was $<0.02$ in both corpora, which are below the standard $0.05$ significance level.

To facilitate the discussion of the ensemble combination situation, the top 10 integrative ensembles are listed in Table 5. The top 10 BERT model combinations used in the integrative ensemble exhibited different trends for Corpus A and Corpus B, as discussed in Subsection 4.2. 

The feature and classifier combinations contained either labels 3 (Ada+Phrase) or lables 6 (RF+Phrase). Char-bigram and token-unigram have some overlapping information. For example, a two-letter token is included in the char-bigram. However, the phrase pattern differs from the aforementioned two features in that it can remove as many textual topics as possible because the content words in a phrase are masked by its POS tag. We believe this property is effective for ensembles. 

The last row of Table 5 summarizes the results of the ensemble for all models. The results showed that the scores of all models were higher than those of the single model, although there was Lower than the highest score. 

\begin{table*}[h]
\centering
\small 
\caption{Top 10 F1 scores for ensemble and integrative ensemble results (1: Ada+Char, 2: Ada+Token, 3: Ada+Phaser, 4: RF+Char, 5: RF+Token, and 6: RF+Phaser)}
\vspace{0.2cm}
\begin{tabular}{ccccccc}
\toprule
 & \multicolumn{2}{c}{BERTs} & \multicolumn{2}{c}{Features and Classifiers} & \multicolumn{2}{c}{Integrated Ensemble} \\ 
 Corpus & Ensemble Labels & F1 & Ensemble Labels & F1 & Ensemble Labels & F1 \\ \midrule
 & \{A, S\} & 0.990 & \{1, 3, 5\} & 0.912 & \{A, S | 3, 5\} & \textbf{1.000} \\ 
 & \{T, A\} & 0.980 & \{1, 3, 4, 5\} & 0.912 & \{A, S | 3, 6\} & \textbf{1.000} \\ 
 & \{A, AW\} & 0.980 & \{1, 3, 4, 5, 6\} & 0.912 & \{A, S | 4, 6\} & \textbf{1.000} \\ 
 & \{A, De\} & 0.980 & \{1, 3, 5, 6\} & 0.901 & \{A, S | 5, 6\} & \textbf{1.000} \\ 
 & \{AW, S\} & 0.980 & \{1, 3, 4\} & 0.893 & \{A, S | 3, 5, 6\} &\textbf{1.000} \\ 
 Coropus A& \{A, AW, De\} & 0.980 & \{1, 2, 3, 6\} & 0.883 & \{A, S | 4, 5, 6\} & \textbf{1.000} \\ 
 & \{T, A, AW\} & 0.970 & \{1, 2, 3, 4, 6\} & 0.883 & \{T, A | 3, 6\} & 0.990 \\ 
 & \{A, AW, S\} & 0.970 & \{1, 2, 3, 5, 6\} & 0.883 & \{T, A | 1, 3, 6\} & 0.990 \\ 
 & \{T, A, AW, De\} & 0.970 & \{1, 2, 3, 4, 5, 6\} & 0.883 & \{T, A | 3, 4, 6\} & 0.990 \\ 
 & \{T, A, AW, S\} & 0.970 & \{1, 3, 6\} & 0.882 & \{T, A | 3, 5, 6\} & 0.990 \\ 
  &All Models & 0.940 & All Models & 0.883 & All Models & 0.990 \\ \midrule
 & \{T, A, AW, De\} & 0.902 & \{1, 2, 6\} & 0.889 & \{T, De | 1, 2\} & \textbf{0.960} \\ 
 & \{A, AW, De, S\} & 0.901 & \{1, 2, 4, 6\} & 0.889 & \{T, AW | 1, 2\} & \textbf{0.960} \\ 
 & \{T, A, De, S\} & 0.894 & \{1, 2, 5, 6\} & 0.889 & \{T, AW | 1, 2, 4\} & \textbf{0.960} \\ 
 & \{T, A, AW, De, S\} & 0.891 & \{1, 2, 4, 5, 6\} & 0.889 & \{T, AW, De | 1, 2, 4, 6\} & \textbf{0.960} \\
 Corpus B & \{T, AW, De, S\} & 0.890 & \{1, 2, 4\} & 0.869 & \{T, AW, De | 1, 2, 4, 5, 6\} & \textbf{0.960} \\ 
 & \{T, A, AW\} & 0.882 & \{4, 5, 6\} & 0.866 & \{AW, De, S | 1, 2\} & \textbf{0.960} \\ 
 & \{T, AW, De\} & 0.882 & \{1, 4, 5, 6\} & 0.859 & \{AW, De, S | 1, 6\} & \textbf{0.960} \\ 
 & \{T, A, AW, S\} & 0.881 & \{1, 2, 5\} & 0.858 & \{AW, De, S | 1, 2, 4\} & \textbf{0.960} \\ 
 & \{A, AW, De\} & 0.880 & \{1, 2, 4, 5\} & 0.858 & \{AW, De, S | 1, 2, 5\} & \textbf{0.960} \\ 
 & \{AW, De\} & 0.880 & \{1, 2, 3, 4\} & 0.855 & \{AW, De, S | 1, 2, 6\} & \textbf{0.960} \\ 
 &All Models & 0.891 & All Models & 0.855 & All Models & 0.950 \\ 
\bottomrule
\end{tabular}
\label{tab:top10_f1_scores}
\end{table*}

\section{Discussion}
Corpus A is included in the pre-training data for Model A (pre-trained on Aozora Bunko) and Model AW (pre-trained on Aozora Bunko and Wikipedia). Therefore, in the integrated ensemble for Corpus A, we considered excluding both models. The highest F1 score for the integrated ensemble, excluding both models, is 0.92 for \{T, De | 1, 3, 5, 6\}, \{T, S | 1, 3, 5, 6\}, and　\{De, S | 1, 3, 5\}. This value is 22.9 and 11 points higher than the highest scores of the BERT models alone (0.691) as well as those of the features and classifiers (0.810), respectively. Additionally, it is 17.6 points higher than the highest score, 0.774, for the ensemble of BERT models \{De, S\}, excluding Models A and AW, and 0.8 points higher than the highest score of the ensemble of features and classifiers (0.912).

For Corpus B, the results for the integrative ensemble are 13.7 and 16 points higher than the highest scores of the BERT model alone (0.823), as well as those of the features and classifiers (0.80), respectively. Additionally, it is 5.8 and 7.1 points higher than the highest scores of the ensemble of BERT models (0.902), as well as the ensemble of features and classifiers (0.889), respectively. Thus, the integrated ensemble significantly outperformed the individual models in both corpora. This result is higher than the ensemble of multiple BERTs and a single feature [39] or a single BERT and multiple classifiers [40].

To further validate the robustness of our integrated ensemble method (I), we conducted Welch's two-sample t-tests comparing it with four baseline ensembles (B, E, G, H). The results demonstrated that I significantly outperformed B, E, and H at $p < 0.001$. In contrast, the I vs. G comparison yielded a relatively higher yet still statistically significant $p$-value ($p < 0.02$, still below the 0.05 threshold), likely attributable to G’s limited sample size (n = 6). Supported by both F1 scores and statistical validation, these findings confirm that performance differences are not due to random chance. The ensemble effect can be attributed to the fact that individual models draw from different aspects of information, enabling them to complement each other when combined in an ensemble. For example, the ensemble score of BERT Models T and A trained on two different sets of pre-trained data was higher than that of BERT Model AW pre-trained on the combination of two datasets. This indicates that pre-trained models are affected not only by the data used during pre-training but also by the various model parameters.

In addition, as analyzed in Subsections 4.3, 4.4, and 4.5, Model S had the lowest score in BERT, and Ada+Phrase and RF+Phrase had the lowest scores yielded among the feature-based models. Nevertheless, they play a significant role in achieving the highest scores for each ensemble and the integrated ensemble. Thus, both individual performance and the nature of the model are important when performing ensembles. 

As shown in the last row of Table 5, the score for the integrated ensemble of all models used was lower than the highest score but was clearly a significant improvement over the highest score for the single model. Compared to the highest score for a single feature and classifier that is not influenced pre-trained data, Corpora A and B improved by 18 and 15 points, respectively. 

Notably, ensembling more models does not necessarily improve performance directly. In some cases, including models with significantly lower scores and weaker characteristics can be counterproductive. This is particularly true when performing weighted ensembles with scores as weights. However, the performance of the integrated ensemble could be further improved by incorporating more diverse, high-performing models. 

The pre-trained BERT model was effective for estimating the authorship of short literary works, even with small samples. The corpus used contained 20 works per author, which is smaller than that of previous studies on text classification using BERT. Nevertheless, BERT tends to score better than conventional feature-based methods, indicating its effectiveness in the AA task. 

In pre-trained models, the pre-training data used for BERT has a non-negligible impact on the task. For example, for Corpus A, the scores of Models A and AW were 27.8 points higher than those of other models. This was because Corpus A was included in the pre-training data. In Corpus B, the scores of Models T, De, and S were 10.2, 13.2, and 9.2 points higher than those in Corpus A, respectively. This was because Corpus A contained pre-1950 works, whereas Corpus B contained post-1990 works and is chronologically closer to the pre-training data for Models T, De, and S.

In this study, the F1 score of 10 author attributions for a Japanese literary work of 510 tokens was more than 0.96. This result is comparable to that of [14] using Corpus A and [15] using Corpus B for feature-based results using full novel texts.

Owing to condition constraints, we used two corpora, five BERTs, three feature sets, and two classifiers. Moreover, we focused on the effectiveness of the integrative ensemble; therefore, we did not test whether the pre-trained BERT models and the features and classifiers used were the best. The score could be further improved by fine-tuning BERT; the proposed method may achieve even better results by incorporating more recent models. Although the ensemble scores varied when constituent models were replaced, the effectiveness of the integrated ensemble remains undeniable.

In this study, 510 tokens from the beginning of each literary work were used. However, indicators of authors’ styles exist throughout the entire work. The estimated score can be improved using more information in attributing authorship to works with more than 510 tokens. The results of the weighted ensemble, using the scores of each model as weights, showed no superiority over the unweighted method. More research should be done on these issues.

\section{Conclusion}
We presented the process and results of a two-corpus study that examines the effectiveness of integrated ensemble PLM- and feature-based approaches in a small-sample AA task. Additionally, we investigated the impact of the pre-training data used for BERT on the performance of this task. The corpora are two sets of self-generated literary works. For the integrated ensemble, we used five BERTs, three types of features, and two classifiers. A summary of the results is presented below.

The score of the proposed integrated ensemble was significantly higher than the highest score of the single model used, confirming its effectiveness. When the corpus that was not included in the pre-training data was used, the integrative ensemble improved the F1 score by approximately 14 points, compared to the highest score single model. Our proposed method achieved the highest score among all approaches tested. Furthermore, BERT is more effective than feature-based in AAs of short literary works, even with small samples, and BERT pre-training data has a significant impact on the task. 

The findings of this study provide useful information not only for AA tasks, but also for general tasks in the development of models, such as PLM and LLM, and in the application of text classification.

\section*{Acknowledgments}
We extend our gratitude to Dr. Yejia Liu for generously permitting us to use Corpus B. Additionally, we appreciate the reviewers for their valuable and constructive feedback.


\section*{Repository}
\begin{itemize}[itemsep=0.15em, parsep=0pt] 
    \item [R1] \url{https://taku910.github.io/mecab/}
    \item [R2] \url{https://clrd.ninjal.ac.jp/unidic/download.html#unidic_bccwj}
    \item [R3] \url{https://taku910.github.io/cabocha/}
    \item [R4] \url{https://huggingface.co/tohoku-nlp/bert-base-japanese-v2}
    \item [R5] \url{https://github.com/akirakubo/bert-japanese-aozora}
    \item [R6] \url{https://huggingface.co/ku-nlp/deberta-v2-base-japanese}
    \item [R7] \url{https://huggingface.co/stockmark/bart-base-japanese-news}
\end{itemize}

\end{document}